\documentclass[conference]{IEEEtran}

\usepackage{cite}
\usepackage{amsmath,amssymb,amsfonts}
\usepackage{graphicx}
\usepackage{textcomp}
\usepackage{xcolor}
\def\BibTeX{{\rm B\kern-.05em{\sc i\kern-.025em b}\kern-.08em
    T\kern-.1667em\lower.7ex\hbox{E}\kern-.125emX}}

\usepackage{algorithm}
\usepackage{array}
\usepackage[caption=false,font=normalsize,labelfont=sf,textfont=sf]{subfig}
\usepackage{stfloats}
\usepackage{verbatim}
\usepackage{hhline}
\usepackage{ragged2e} 
\usepackage{enumitem} 
\usepackage{multirow}%
\usepackage{booktabs}%
\usepackage{algorithmicx}%
\usepackage{algpseudocode}%
\usepackage{listings}%

\algdef{SE}[SUBALG]{Indent}{EndIndent}{}{\algorithmicend\ }%
\algtext*{Indent}
\algtext*{EndIndent}

\usepackage{adjustbox}
\usepackage{array}

\newcolumntype{R}[2]{%
	>{\adjustbox{angle=#1,lap=\width-(#2)}\bgroup}%
	l%
	<{\egroup}%
}

\begin{document}

\title{QBSD: Quartile-Based Seasonality Decomposition for Cost-Effective RAN KPI Forecasting}


\author{\IEEEauthorblockN{Ebenezer~R.H.P.~Isaac and Bulbul Singh}
	\IEEEauthorblockA{\textit{Global AI Accelerator, Ericsson},
		Chennai, India \\
		ebenezer.isaac@ericsson.com, bulbul17110@gmail.com}}

\maketitle

\begin{abstract}
	Forecasting time series patterns, such as cell key performance indicators (KPIs) of radio access networks (RAN), plays a vital role in enhancing service quality and operational efficiency. State-of-the-art forecasting approaches prioritize accuracy at the expense of computational performance, rendering them less suitable for data-intensive applications encompassing systems with a multitude of time series variables. They also do not capture the effect of dynamic operating ranges that vary with time. To address this issue, we introduce QBSD, a live single-step forecasting approach tailored to optimize the trade-off between accuracy and computational complexity. The method has shown significant success with our real network RAN KPI datasets of over several thousand cells. In this article, we showcase the performance of QBSD in comparison to other forecasting approaches on a dataset we have made publicly available. The results demonstrate that the proposed method excels in runtime efficiency compared to the leading algorithms available while maintaining competitive forecast accuracy that rivals neural forecasting methods.
\end{abstract}


\section{Introduction}
\label{sec:intro}

Recent advancements in machine learning (ML) have led to the development of powerful and flexible time series forecasting techniques based on deep learning (DL)~\cite{lim2021time} and ensemble methods \cite{allende2017ensemble} especially for long-horizon forecasting. The landscape of DL-based forecasting methods includes transformers \cite{zhou2021informer}, recurrent neural networks (RNNs) \cite{HEWAMALAGE2021388} and various combinations of MLP (multi-layer perception) and convolutional neural network (CNN) \cite{borovykh2017conditional} structures.
These models have shown significant improvements in forecasting accuracy over traditional statistical models, such as as ARIMA~\cite{Harvey1990} and SARIMA~\cite{SARIMApaper}, by capturing complex temporal dependencies and nonlinear patterns in the data. Due to their prevalence in literature, the forecasting approaches that involve any form of neural network are categorized as ``neural forecasting methods." 
Despite the existence of numerous forecasting methods, ensuring both accuracy and computational efficiency remains a challenge, particularly when dealing with anomalies or irregularities in time series data.  
Many existing time series forecasting methods are computationally intensive hindering their practicality for large-scale application. Moreover, several forecasting methods necessitate complex hyperparameter tuning. For instance, even a slight deviation in any of the four parameters of SARIMA could lead to an increased forecasting error.

Though neural forecasting methods are considered superior in recent literature, they are impractical for scenarios involving a system consisting of a large number of time series variables with a need for frequent retraining. Such cases are common in RAN (radio access network) applications, e.g., performance management (PM) KPI anomaly detection, sleeping cell detection, cell traffic forecasting and load balancing. A standard cell in RAN can generate over 300 network PM KPIs every 15 minutes. A typical RAN topology is composed of several thousand such cells. Depending on the use case, the solution can be expected to forecast over 300,000 time series variables every 15 minutes. Furthermore, forecasting is only part of the solution for the aforementioned use cases; the system needs to apply additional computation on top of the forecasted outcomes. Due to the dynamic nature of these KPIs and the possibility of drifts, the solution would also require periodic retraining. From a business perspective, there would be no financial gain in employing hundreds of dedicated GPUs/CPUs to work continuously for the use cases concerned, making the viability of neural forecasting solutions questionable for such applications. Therefore, there is a compelling need for a computationally efficient forecasting approach that has comparable forecasting accuracy to neural forecasting methods.

Research in predictive analytics often emphasizes developing long-horizon forecasting methods, potentially underestimating the advantages and applications of single-step forecasting techniques. Single-step forecasting methods are preferable for real-time decision-making, handling large-scale volatile data, and situations requiring frequent updates with limited computational resources.
\IEEEpubidadjcol

This paper proposes QBSD (Quartile-Based Seasonality Decomposition, filed as a US by Ericsson \cite{QBSDpatent})
to address the need for a computationally efficient network KPI forecasting approach.  It is designed for rolling single-step forecasts on live data and does not require a separate fit and predict stage. This technique was mentioned in \cite{ath-paper} for data preparation prior to time series anomaly detection, without describing QBSD's algorithm and its implementation. In this paper, we delve into such details and show how it compares against several state-of-the-art and popular forecasting methods. The evaluation reveals that the proposed method is superior in runtime efficiency compared to the best available algorithms while still being competitive in forecast accuracy as indicated by MAPE (mean absolute percentage error).

The following are the significant contributions of this article:
\begin{enumerate}
	\itemsep2pt
	\item A computationally efficient rolling forecasting algorithm with only two simple hyperparameters that follows a statistical approach with an accuracy that rivals neural forecasting approaches. 
	\item  Estimation of operating ranges that vary with time. These bounds are exposed for interpretability and business decision making.
	\item Evaluation of the proposed work with the state of the art on publicly available datasets comparing both forecasting accuracy and execution time.
	\item Evaluation on a network KPI dataset to further emphasise the applicability of the proposed method for RAN KPI use cases, and enable further research progression by making the dataset publicly available.
\end{enumerate}

\section{Related work}
\label{sec:relatedwork}

The spectrum of time series forecasting literature can be broadly divided into statistical approaches, neural forecasting, hybrid models and other ML-based techniques. This section reviews the literature along with their relative strengths and limitations.

\subsection{Statistical Approaches}
Traditional statistical models such as Autoregressive Integrated Moving Average (ARIMA) \cite{Harvey1990} and Seasonal ARIMA (SARIMA) \cite{SARIMA} has been used widely in various domains such as forecasting the demand in the food industry \cite{fattah2018forecasting}, and have demonstrated promising performance in capturing linear trends and seasonality patterns. One of the most successful statistical models is Facebook (FB) Prophet \cite{taylor2018forecasting} with the ability to incorporate the effect of holidays in business-level forecasting applications. Nevertheless, statistical models often struggle with non-linear and complex patterns, limiting their effectiveness in specific scenarios such as long-horizon forecasting. However, due to their simplicity and computational efficiency, they are still preferred over deep learning methods today for big data applications \cite{forecastingComparison}.

\subsection{Neural Forecasting}
The DeepAR proposed by Salinas et al. \cite{SALINAS20201181} is based on an autoregressive RNN that can model the distribution of future values given past observations. 
NeuralProphet \cite{NeuralProphet} uses a variant of RNN to capture temporal dependencies in time series data and provides uncertainty estimates for the forecasts. The authors claim that NeuralProphet outperforms Prophet and is capable of handling large-scale datasets. 

The transformer architecture has been adapted for time series forecasting to yield competitive results. Lim et al. \cite{LIM20211748} proposed a Temporal Fusion Transformer (TFT), an attention-based design that combines high-performance multi-horizon forecasting with comprehensible insights into temporal dynamics. 
The Crossformer \cite{zhang2022crossformer} is a transformer-based architecture that captures cross-dimension dependency using dimension-segment-wise embedding, a two-stage attention layer and a hierarchical encoder-decoder.

N-BEATS (Neural Basis Expansion Analysis for interpretable Time Series forecasting) \cite{oreshkin2019n} is a popular deep neural architecture based on backward and forward residual links with multiple stacks of MLP constructs (multi-layer FC) with ReLU nonlinearities. N-HiTS \cite{challu2022nhits} is a recent improvement over N-BEATS through multi-rate sampling and multi-scale hierarchical interpolation but retaining the base stacked MLP structure.


Regardless of the type of neural approach adopted (old or new), DL models generally require extensive computational resources and longer training times limiting their practical applicability for large-scale production applications with resource constraints. Though successive models claim to be more computationally efficient than their predecessors, DL models cannot be as efficient as their statistical counterparts. 

\subsection{Hybrid Models}
Hybrid models aim to combine the strengths of multiple forecasting techniques to improve overall performance. 
Spranger et al. \cite{SPRANGERS2023332} proposed a bidirectional temporal convolutional network (BiTCN) that requires fewer parameters than the conventional Transformer-based approach. The study shows that BiTCN is more computationally efficient than the commonly used bidirectional LSTM. 
Zheng et al. \cite{ZHENG202393} compared the performance of recent hybrid models for traffic prediction. The study concluded that the parallelized architecture outperforms the stacked architecture, and models that can extract dynamic spatial features outperform models that focus solely on dynamic temporal feature analysis.  

The drawback of hybrid models is the increased complexity in model design and parameter tuning. Integrating multiple models requires careful selection and optimization of individual model components, alongside determining the optimal weighting scheme for combining their predictions. Additionally, hybrid models may introduce additional computational overhead as combining different algorithms requires more computational resources than single-model approaches. While there are approaches like BiTCN that consider reducing the computational overhead, studies like \cite{du2022bayesian} forego the consideration of computational complexity altogether by building a framework to learn a weighted combination of multiple statistical, ML-based and neural forecasters.

\subsection{Other ML Techniques}

Berry et al. \cite{BERRY2020552} proposed a Bayesian approach to estimate consumer sales. 
The study arose from the realization that the variability observed in high-frequency sales is due to the compounding impact of variability in the number of transactions and sales-per-transaction. BayesMAR is a simple strategy of extending the traditional AR model to a median AR model (MAR) for time series forecasting is proposed in \cite{ZENG20211000}. 

In \cite{JANUSCHOWSKI20221473} Januschowski et al. described why tree-based methods were so popular in the M5 competition. The paper discussed software packages that employ gradient boosting models, including LightGBM \cite{NIPS2017_6449f44a} and XGBoost \cite{XGBoost}, which have demonstrated robustness and high performance. 
Tree-based models can be preferred over DL models for data-intensive applications since they are far superior in computational efficiency. However, tree-based models require careful tuning of hyperparameters as they may be prone to overfitting, especially when dealing with noisy or sparse data.

\subsection{The Gap}
In summary, several strategies, ranging from conventional statistical models to more modern ML-based techniques, have been presented for time series forecasting. 
While existing forecasting methods have progressed in tackling various challenges, some issues remain unresolved. These issues include the requirement for complicated hyperparameter tuning and separate fit and prediction stages in existing methods which result in high computational costs. Literature has yet to cover the aspect of the operating range of values that varies with time. This phenomenon is mostly seen in telecom data wherein the standard deviation of the data is higher during the day and lower close to midnight. 

\begin{figure*}[t]
	\centering
	\vspace{1em}
	\includegraphics[width=0.9\textwidth,height=\textheight,keepaspectratio]{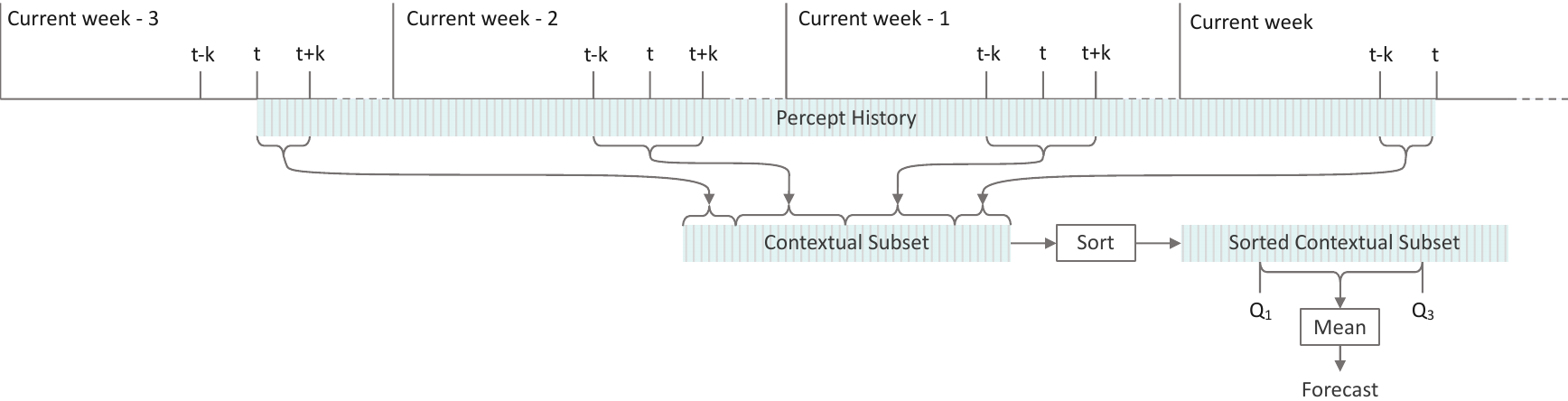}
	\caption{Schematic architecture of QBSD forecast computation, where $t$ is the current timestamp and $k$ is the context period.}
	\label{fig:Contextual_subset}
\end{figure*}

\section{Method}
\label{sec:method}


 The target architecture is based on the following criteria:
 \begin{itemize}
 \itemsep0pt
  \item Minimal parameters required for functioning
  \item Avoidance of complicated hyperparameter tuning
  \item Estimation of operating ranges that vary with time
  \item Solution that can be generalized to a wide spectrum of cell traffic KPIs
  \item Rolling forecast model that does not require a scheduled retraining
  \item Prioritize computational efficiency and interpretability while retaining a forecasting accuracy that can compete against neural forecasting approaches
 \end{itemize}

 \begin{algorithm}[t]
	 \caption{Quartile-Based Seasonality Decomposition}
	 \small
	 \begin{algorithmic}
		 \Require Timestamp $t$, time series $M$, context period size $k$
		 \State Compile contextual subset $S$ from history (total of $6k+3$ samples)
		 \Indent
		 \State $M(t-k)$ through $M(t-1)$ for $\mathit{day}(t)$ of $\mathit{week}(t)$
		 \State $M(t-k)$ through $M(t+k)$ for $\mathit{day}(t)$ of $\mathit{week}(t)-1$ and $\mathit{week}(t)-2$
		 \State $M(t)$ through $M(t+k)$ for $\mathit{day}(t)$ of $\mathit{week}(t)-3$
		 \EndIndent
		 \State Calculate quartiles $Q_1$ and $Q_3$ of $S$ 
		 \State $\mathit{IQR} = Q_3 – Q_1$
		 \State $\mathit{forecast}(t)\; =\; \text{mean}({ x \mid x \in S\; \text{and} \; Q_1<x<Q_3})$
		 \State $\mathit{diff\textunderscore residual} (M,t)\;=\; M(t)-\mathit{forecast}(t)$
		 \State $\mathit{norm\textunderscore residual} (M,t)\;=\;  (M(t)-\mathit{forecast}(t))/\text{max}(\mathit{IQR}(t), c)$ \\
		 \Return $Q_1, Q_3, \mathit{IQR}, \mathit{forecast}$, residuals
	\end{algorithmic}
	\label{alg}
\end{algorithm}

\subsection{QBSD Algorithm}

The basic version of QBSD addresses daily and weekly seasonality by assessing the historical patterns of the past month. The system maintains a first-in, first-out (FIFO) buffer containing a rolling 4-week window of time series data, updated with each data point captured in the stream. Only a subset of the context window holds the most relevant information to forecast the value of the next timestamp. This subset is called the \textit{contextual subset}. The contextual subset is built based on the following assumptions: (1) At any given time of the day, the values corresponding to timestamps closest to the timestamp to be forecasted provide a better contribution to estimating the forecast value than the other timestamps. (2) The data distribution on a given day of the week is similar to the same day of the week for the most recent weeks. The contextual subset compilation and forecasting operation is succinctly illustrated in Fig.~\ref{fig:Contextual_subset}. 

Let $t$ denote the timestamp for which the forecasted value is to be determined, $k$ represent the \textit{context period}, indicating the number of timestamps closest to $t$ that are to be taken into consideration, $\mathit{day}(t)$ be the day of week corresponding to $t$, $\mathit{week}(t)$ be the week of the year corresponding to $t$, $M(t)$ symbolize the value of series $M$ at $t$, and $S$ denote the contextual subset.

 \begin{figure*}[t]
     \centering
     \includegraphics[width=0.85\textwidth,keepaspectratio]{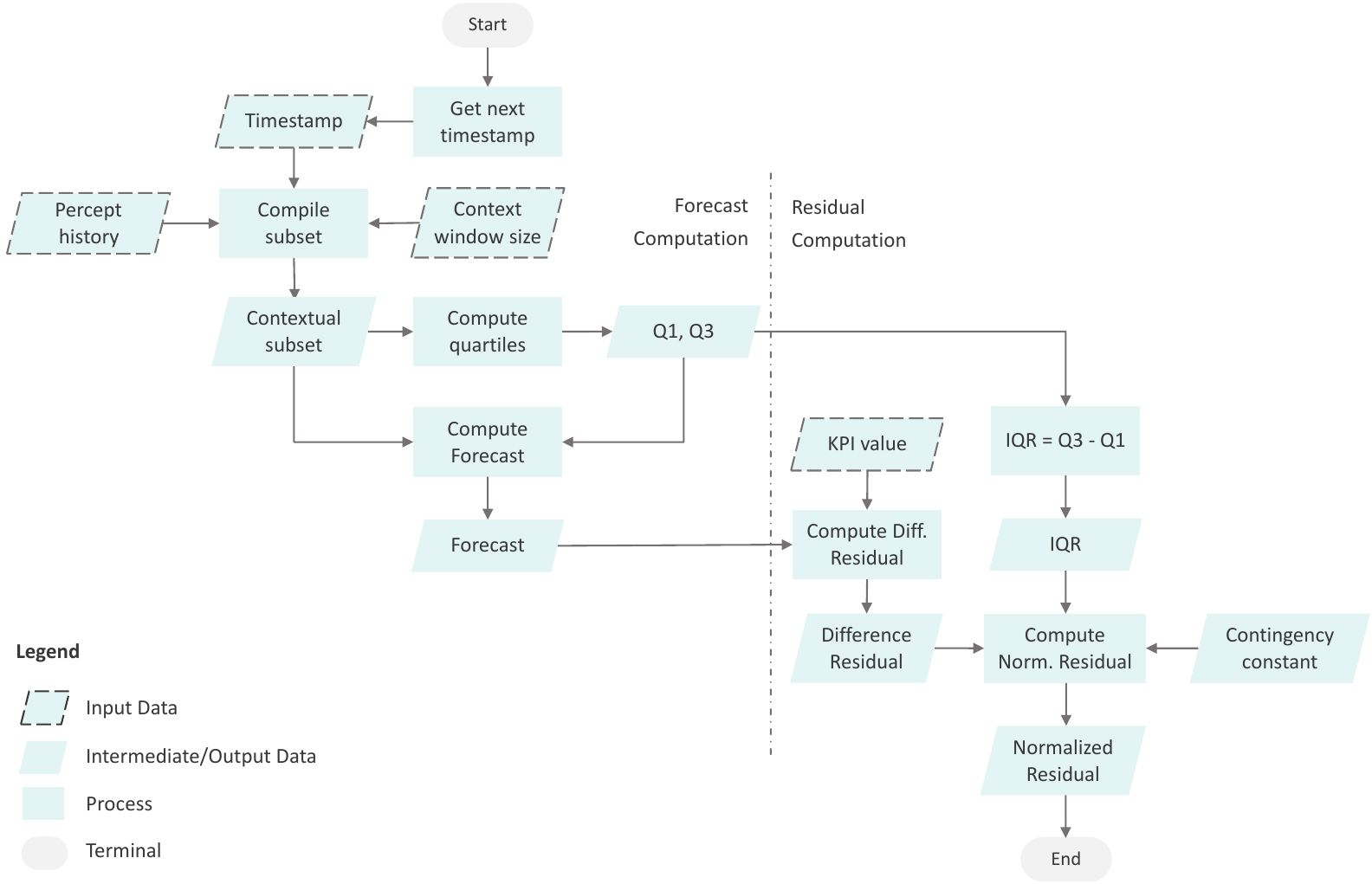}
     \caption{Flowchart illustrating QBSD. The flow begins with the input timestamp for which the forecast should be generated. This forecast along with the observed KPI value of the given timestamp is used for residual computation.}
     \label{fig:QBSD_Flowchart}
 \end{figure*}

$S$ is compiled as follows: take the values $M(t-k)$ through $M(t-1)$ for the current day, i.e., $\mathit{day}(t)$ in $\mathit{week}(t)$. Then take values from $M(t-k)$ through $M(t+k)$ of the same day for the past two weeks, $\mathit{day}(t)$ of $\mathit{week}(t)-1$ and $\mathit{week}(t)-2$. Finally, take $M(t)$ through $M(t+k)$ of $\mathit{day}(t)$ of $\mathit{week}(t)-3$. The quartiles, $Q_1$ and $Q_3$, are computed from $S$. The mean of the values that lie between these quartiles within $S$ is the forecast for timestamp, $t$. The interquartile range, $\mathit{IQR} = Q_3 – Q_1$, becomes the expected deviation for timestamp, $t$

The quartile values (and hence the resulting $\mathit{IQR}$) are different for each timestamp, hence deriving the expected operating range for each timestamp at any given time of the day; $Q_1$ depict the expected lower range whereas $Q_3$ depict the expected upper range. While the algorithm samples all values from time $t-k$ up to $t+k$ for the past two weeks, it only considers $t$ to $t+k$ for the third week. This is because, in the current week, it is only possible to take values from time $t-k$ up to $t-1$. The sampling is done so that no period would have an additional overlap bias while calculating the quartiles. Note that the contextual subset is built based on elements within temporal intervals and not array indices. This adds flexibility to the algorithm to cope with missing values. As long as there is sufficient number of elements within the contextual subset, missing value imputation will not be necessary.  

The algorithm computes two types of residuals: the difference residual, which is the difference between the forecasted and actual values, and the normalized residual, obtained by dividing the difference residual by $\text{max}(\mathit{IQR}, c)$, where $c$ is the \textit{contingency constant} for each $t$. The normalized residual adjusts for the dynamic operating range and mitigates sensitivity to lower ranges by including $c$. This constant not only prevents division by zero when $\mathit{IQR}=0$, but also adjusts for the practical significance of deviations in the data, such as minor peaks in time series data that should not be flagged as anomalies in certain contexts. The choice of $c$ thus balances statistical rigor with practical considerations and can be tailored to the application's needs, ranging from a nominal value (like 1) for sensitive applications to a value derived from the 1-percentile of the training dataset for general cases. For example, typical values of Active Uplink Users KPI usually range in the order of thousands for a densely-populated urban cell; during midnight, it is almost always 0. Hence, even a minor peak, such as 5, is considered a significant outlier in a statistical sense. However, when the operating range is so high, such a minor magnitude should not be considered an anomaly in a practical scenario. So, in this example, the value of $c$ can be between 10 to 100 to avoid unnatural spikes in the normalized residual arising from nominal fluctuations in the time series data at lower expected ranges. 

The flowchart for the QBSD method is depicted in Fig.~\ref{fig:QBSD_Flowchart}. This chart includes the steps involved to generate the expected bounds (quartiles), forecast and residuals for a single timestamp. This process is repeated for every subsequent timestamp in production.

The proposed method implicitly handles seasonal variations, long-term secular trends and cyclic fluctuations to obtain irregular variations, but it does not consider short-term trends. 
A sudden change in trend could be potentially anomalous must be captured (not decomposed) when it comes to RAN KPIs to make appropriate business decisions.

The 4-week window size was selected based on empirical evaluation to minimize forecasting error. If the size is too small, then the weekly seasonality will not be captured effectively. If the size is too big, the system will be slow to respond to change in data drifts. While the basic version of QBSD accounts for daily and weekly seasonality, the algorithm can be altered to capture monthly and yearly seasonality as well depending on the granularity of the data and the context window size. If the interval between each successive data point of the given dataset is one day, then the context period will be in the order of days and the context subset can be compiled by spanning between months or years depending on the requirement of the use case. Quartiles are used to build contextual subsets as reasonable defaults to remove the effects of outliers caused by the stochastic nature of network KPIs. Broader quantile ranges increase sensitivity to outliers and narrower quantile ranges leads to losing valuable information in the series. 

Since both the quartile series can be used as a visual aid for the operator to make informed decisions, these series can be smoothed before being plotted. The smoothing method that can be employed is a solution design choice. For example, it can be a Savitzky-Golay filter \cite{savitzky1964smoothing} or even a simple moving average. 

\subsection{Example}

\begin{figure*}
	\centering
	\includegraphics[width=0.9\linewidth,height=\textheight,keepaspectratio]{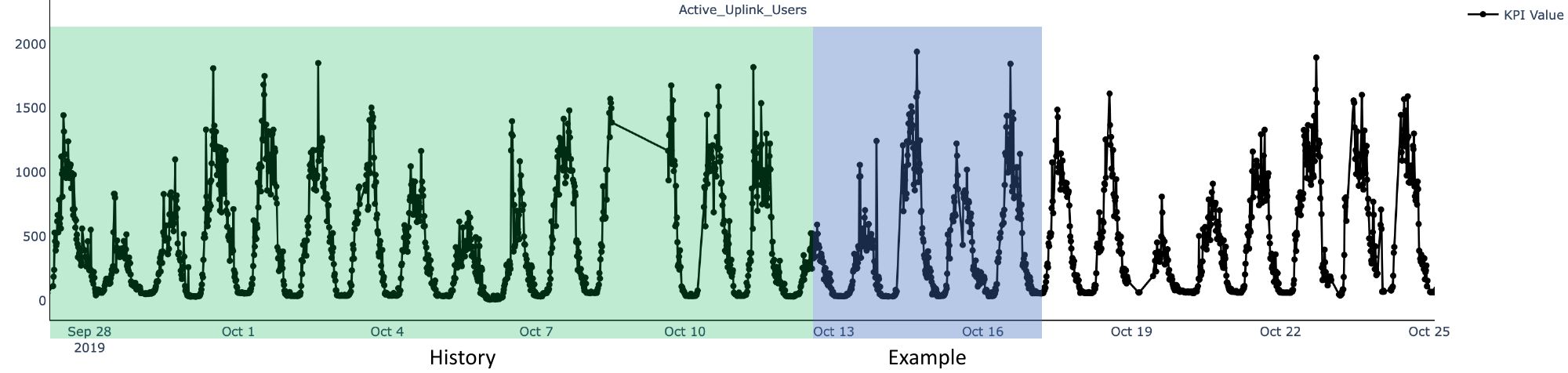}
	\caption{Sample input data. Percept history is highlighted in green while QBSD is performed on the example sequence highlighted in blue. Note that there can be missing data in history.}
	\label{fig:sampleinput}
\end{figure*}

\begin{figure*}[b]
	\centering
	\includegraphics[width=0.9\linewidth,height=\textheight,keepaspectratio]{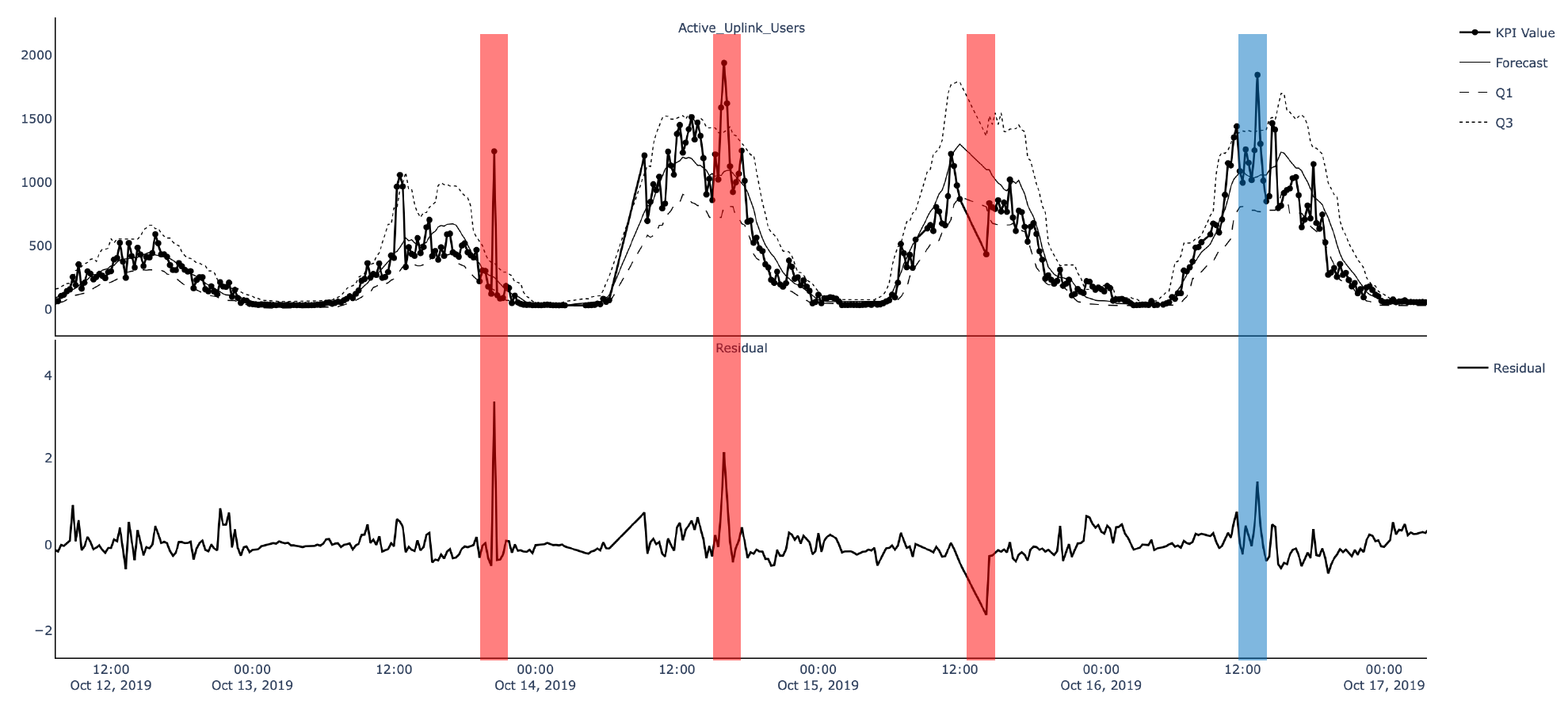}
	\caption{Illustration of QBSD applied on the example highlighted in Fig.~\ref{fig:sampleinput} along with normalized residuals. Red bars indicate definitive anomalies wherein the blue bar indicates a peak that is not statistically an anomaly.}
	\label{fig:illustration}
\end{figure*}

This section illustrates a short example of the QBSD application to a simple anomaly detection use case. Fig.~\ref{fig:sampleinput} includes a snapshot of Active Uplink Users KPI over two months; only four weeks are shown here for clarity. Note how the data follows both daily and weekly seasonal patterns. The cyclic pattern that occurs every day depicts daily seasonality. The periodic pattern of five consecutive daily peaks (corresponding to weekdays) followed by two smaller peaks (corresponding to weekends) depict weekly seasonality. Also note that this dataset contains missing data, e.g., between Oct 8, 12:30 and Oct 9, 15:45. Nevertheless, there were enough samples for the previous weeks for the same period to accurately estimate the forecast and range for that period. This data was passed through the QBSD algorithm to illustrate the seasonality decomposition. For this example, the context period was $k= 2.5$ hours and the contingency constant was $c=1$. 

Figure~\ref{fig:illustration} shows the sample output of the algorithm. The output compares the QBSD forecast with the observed KPI data along with the generated normalized residual. The sample output shows that the proposed method has captured the lower and upper bound effectively. The variation between the expected lower and upper limits differed  throughout the day. The range appeared narrower around midnight and broader during the day. The forecast also adequately approximated the expected value of the KPI. The general notion is that any value that deviates significantly beyond these bounds is probably an anomaly. In this case, the first two spikes and one dip qualified as anomalies.
Not all KPI values that exceed the expected bounds can be considered anomalous. For instance, consider the third spike in this example: it was just above the expected range, but the normalized residual shows that the peak was not significantly greater than the others calculated in recent history. Hence, statistically, the third spike is not an anomaly, but this definition could differ from one use case to another based on the business requirements. 

The severity of an anomaly in this example is inversely proportional to the expected range at the time of deviation. E.g., a moderate deviation at midnight can be considered anomalous while during the day, the deviation ought to be significantly large for it to be alerted as an anomaly.

\subsection{Computational Complexity}
The computational complexity of QBSD per forecast is derived by considering the number of elements in the input data used to compute the forecast. Let $n$ be the number of elements in history fed to the algorithm. Only a subset consisting of $m$ elements from the set of $n$ elements is considered for computation (contextual subset), and the size of that subset must be far lesser than $n$ ($m << n/2$). Hence, $O(\log n)$ can represent the subset of elements that proceed to the next step. The quartile computation involves sorting by Quicksort, followed by subsetting and mean calculation for the forecast resulting in an average of $O(m \log m)$ time. Because $O(m)$ equates to $O(\log n)$, one can regard the resulting time complexity of QBSD as $O(\log (n + \log n))$.

\begin{figure*}[t]
	\centering \includegraphics[width=\textwidth,keepaspectratio]{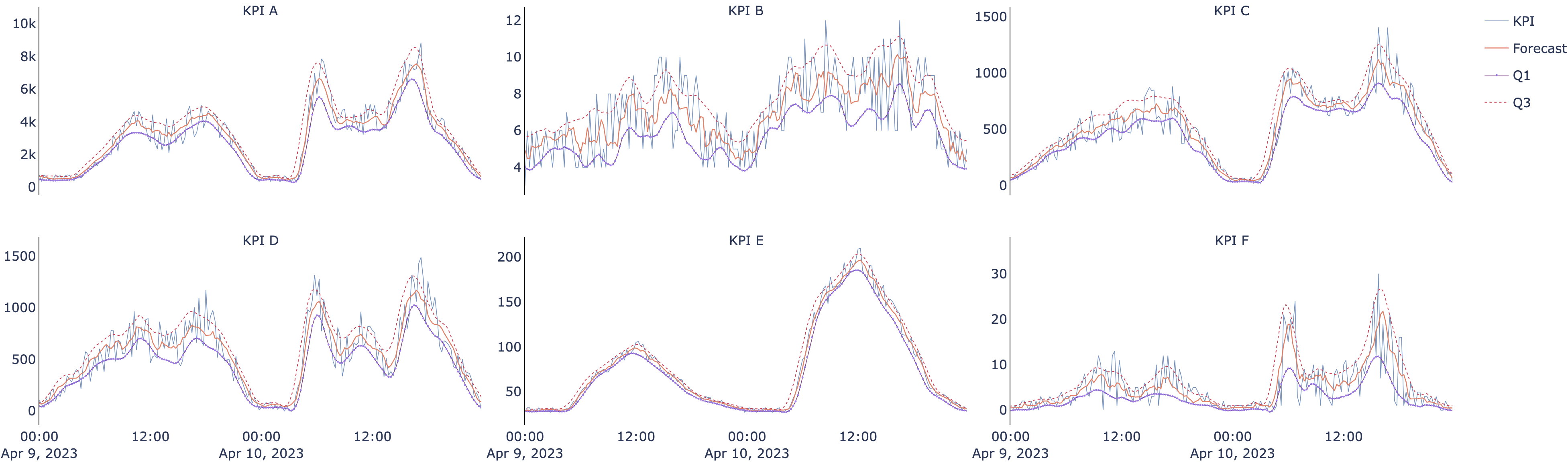}
	\caption{Plot of actual KPI values (A through F) along with forecast, $Q_1$, and $Q_3$ obtained using the QBSD algorithm for the Cell-F dataset. Series $Q_1$ and $Q_3$ have been smoothed with the Savitzky–Golay filter.}
	\label{fig:EON1-QBSD}
\end{figure*}

\section{Experimental setup}
\label{sec:expsetup}
	\subsection{Dataset}
	As we cannot disclose assessments from the real customer network data, we have curated 
	a synthetic dataset, Cell-F
	for univariate time series forecasting 
	from the Ericsson Outlier Nexus (EON) 
	compiled to closely represent the characteristics of real cell PM KPIs.
	The dataset, EON1-Cell-F \cite{eon1} is made publicly available by Ericsson Global AI Accelerator (GAIA) through Ericsson Research.
	Cell-F contains the data of 6 KPIs from February 2023 to April 2023 with a 15-minute interval between data points. The interval is considered the ROP (result output period). Thus, there are 96 ROPs in a day. A moving training window spanning the previous month is employed to train the forecaster that predicts the next timestamp, without considering any predicted values from previous timestamps. This process is followed to prediect each timestamp of one month (from 01-04-2023 00:00:00 to 30-04-2023 23:45:00).
	
	\subsection{Baseline methods for comparison}
	The performance of QBSD is compared with some state-of-the-art and popular forecasting methods. The summary of the methods is shown in Table~\ref{tab:methods}. Since QBSD is a single-step live forecasting algorithm, we have also included a simple baseline of $M(t-1)$, i.e., value of the previous timestamp, to the comparison.
	
	\begin{table}
		\addtolength{\tabcolsep}{-2pt}
		\caption{Methods included in the evaluation}
		\centering
				\begin{tabular}{lllc}
					\toprule
					Method & Reference & Approach & Year \\
					\midrule
					SARIMA & Guin \cite{SARIMApaper} & Statistical & 2006 \\
					XGBoost & Chen and Guestrin \cite{XGBoost} & Tree-based & 2016 \\
					LightGBM & Guolin Ke et al. \cite{NIPS2017_6449f44a} & Tree-based & 2017 \\
					Prophet & Taylor and Letham \cite{taylor2018forecasting} & Statistical & 2018 \\
					N-BEATS & Oreshkin et al. \cite{oreshkin2019n} & Neural & 2019 \\
					N-HiTS & Challu et al. \cite{challu2022nhits} & Neural & 2022 \\
					QBSD & Proposed & Statistical & - \\
					\bottomrule
				\end{tabular}
				\label{tab:methods}
			\end{table}
	
	 \subsubsection{Statistical models}
	 SARIMA \cite{SARIMA} is an extension of the ARIMA model to capture seasonal patterns in the data. Prophet \cite{taylor2018forecasting} is a model developed by Facebook's Core Data Science team that utilizes a decomposable time series model with three main components: growth, seasonality, and holidays. While these methodologies do not pertain to recent innovations, their enduring relevance persists in addressing contemporary business complexities and scholarly advancements. Furthermore, given that QBSD is also a statistical forecasting model, these methods inherently engage in direct juxtaposition with QBSD.
	
	 \subsubsection{Tree-based models}
	 XGBoost \cite{XGBoost} is a gradient-boosting algorithm that combines multiple decision trees to generate forecasts. XGBoost uses a range of optimization techniques to minimize the loss function, including regularization to prevent overfitting and early stopping to avoid unnecessary computation. LightGBM (Light Gradient Boosting Machine)  \cite{NIPS2017_6449f44a} is designed to be faster and more memory efficient than XGBoost. It employs histogram-based gradient boosting in which data is bucketed into bins using a histogram of the distribution. The bins are used to iterate, calculate the gain, and split the data. It has also gained popularity in the M5 competition. 
	
	\begin{table}
		\caption{MAPE comparison on Cell-F}
		\addtolength{\tabcolsep}{-3pt}
		\centering
			\begin{tabular}{lrrrrrrrc}\toprule
				&\multicolumn{6}{c}{Datasets} \\
				\cmidrule{2-7}
				Method &
				A &
				B &
				C &
				D &
				E &
				F &
				Mean & p-value \\
				\midrule
				$M(t-1)$ & 22.23 & 23.42 & 24.98 & 54.09 & 7.61 & 99.32 & 38.61 & \textbf{0.0156} \\
				SARIMA   & 19.41 & 19.91 & 21.91 & 39.82 & 6.88  & 83.60  & 31.92 & 0.1094 \\
				XGBoost  & \textbf{5.46}  & \textbf{6.60}  & \textbf{5.48}  & \textbf{7.16}  & \textbf{2.14}  & \textbf{13.07}  & \textbf{6.65} & 1.0000 \\
				LightGBM & 14.92 & 12.01 & 13.69 & 38.60 & 4.79  & 59.80  & 23.97 & 1.0000 \\
				Prophet  & 39.84 & 19.32 & 35.41 & 75.77 & 26.01 & 116.84 & 52.20 & \textbf{0.0156} \\
				N-BEATS  & 15.41 & 18.38 & 18.14 & 39.90 & 5.74  & 75.42  & 28.83 & 0.8398 \\
				N-HiTS   & 20.57 & 21.87 & 31.54 & 52.00 & 7.84  & 108.34 & 40.36 & \textbf{0.0156 }\\
				QBSD     & 15.70 & 18.89 & 17.78 & 42.08 & 5.14  & 81.88  & 30.25 & -\\
				\bottomrule
			\end{tabular}
		\label{tab:mape2}
		
	\end{table}

	 \subsubsection{Neural forecasting models}
	 N-BEATS \cite{oreshkin2019n} is based on a stack of fully connected layers that learn to decompose a time series into a set of basic functions, which are then used to generate forecasts. N-BEATS has achieved state-of-the-art results in several time series forecasting benchmarks, including the M4 competition. N-HiTS \cite{challu2022nhits} is an improvement over the N-BEATS architecture through multi-rate sampling and multi-scale hierarchical interpolation.
	
	To predict individual data points within the test set, each algorithm in this study utilized a moving training window, akin to the technique employed in QBSD. 
	Prophet was utilized with its default hyperparameters.
	For SARIMA, the optimal values for p, q, and seasonal components P, Q, and m were determined using the auto ARIMA function \cite{autoARIMA}. N-BEATS and N-HiTS were implemented using the darts package \cite{dartsPaper}. 
	For QBSD, the context period was set to $k=1$ hour to compile the contextual subset using values from the current and the previous 3 weeks. 
	
	Though this method is tailored for network cell traffic KPIs, depending on the application, it can be modified for other time series has well. For completeness, interested readers can refer Appendix~\ref{sec:extra} for experimentation on other popular datasets that used to compare forcasting algorithms.
	
	\section{Results and Discussion}
	\label{sec:resultsanddiscussions}
		
	
	\subsection{Statistical Significance}
	The Wilcoxon signed-rank test \cite{conover1971practical} was preferred over the paired t-test for this scenario since the distribution of the difference between the pairs were not normal. The paired t-test's underlying assumption of a normally distributed difference \cite{mcdonald2014handbook} was deemed inapplicable in this context. The p-value signifies the probability of the null hypothesis being true. At a significance level of 5\%, a p-value below 0.05 is deemed significant to warrant the rejection of the null hypothesis in favour of the alternative hypothesis.
	
	In Table~\ref{tab:mape2}, the null hypothesis proposes that QBSD's MAPE was equivalent to or exceeds that of the method undergoing comparison. The alternative hypothesis proposes that QBSD demonstrates a MAPE that was significantly lesser than the method being subjected to comparison. 
		
	\subsection{Forecasting Accuracy}
	
	QBSD performed similarly to the other best-performing algorithms in accuracy for KPIs A, D, and E. Daily and weekly seasonality were observed in all the presented KPIs, with lower peaks during the weekends due to decreased user activity. Fig.~\ref{fig:EON1-QBSD} illustrates the ability of QBSD to capture the lower and upper regions of the data for all the KPIs. The bounds $Q_1$ and $Q_3$ have been smoothed using the Savitzky-Golay filter \cite{savitzky1964smoothing} (SciPy implementation \cite{2020SciPy-NMeth}) for visual aid. Despite the erratic nature of KPI-F, which generated the highest forecasting error, QBSD performed reasonably well in capturing the lower and upper boundaries. For the remaining KPIs, QBSD consistently delivered competitive results. 
	
	QBSD demonstrates an overall MAPE comparable to SARIMA while exhibiting a statistically significant performance over Prophet. In specific instances, QBSD outperforms SARIMA. In neural forecasting, QBSD almost parallels N-BEATS in error margins and consistently achieves a statistically better MAPE compared to N-HiTS across all KPIs. It is important to acknowledge that the performance exhibited by neural forecasting methodologies in this experiment might not comprehensively reflect their overall capabilities. Though the test single-step forecasting, the neural forecasting methods are aimed at long-horizon forecasting which is beyond the scope of this study. Both tree-based approaches notably outperform QBSD in terms of MAPE. According to the evaluation, XGBoost was arguably the method with the least error among all the methods considered in this experiment. For instance, QBSD has a much greater error margin for KPIs D and F compared to XGBoost. Nevertheless, XGBoost lacks the explainability and visibility to operational ranges that is provided by QBSD in addition to its computational efficiency. Through QBSD, one can identify the expected norm and its range for any given period which is not possible in any other method described in this paper.
	
	\subsection{Resource Efficiency}
	In a practical scenario, for example, a cellular network composed of at least 3500 cells, each cell is characterized by at least 100 KPIs (and even goes above 300 KPIs). In such a system, if anomaly detection is to be applied, there needs to be context-sensitive models deployed for each of these KPIs for each cell. This setup leads to deploying at least 350,000 models that should do both forecasting and decomposition in addition to anomaly detection. The computational workload would be immense to consider models such as N-BEATS, XGBoost or LightGBM. The empirical evaluations for runtimes are shared in the Appendix (Table~\ref{tab:runtime}). Not only do these methods incur considerable computational costs, but they also have a much higher space complexity compared to QBSD. 
	The marginal benefits the algorithms provide in accuracy cannot satisfy the additional cost in computational complexity. Therefore, in a real-world application, one must make a fair trade-off in accuracy-based metrics for a performance gain in cost. For instance, \cite{ath-paper} shows how the accuracy provided by QBSD is sufficient to provide cleaner residuals for reliable detection of network anomalies.
		

	\section{Conclusion}
	\label{sec:conclusion}
	This paper describes QBSD, a computationally effective solution to univariate time series single-step forecasting specialized for RAN KPIs that exhibit seasonality. 
	Though QBSD may not be the best method in terms of forecasting accuracy, it excels in speed while retaining a competitive edge on par with neural forecasting methods. It also produces the time-sensitive upper and lower operating ranges along with the forecast. 
	Possible future work can be extending QBSD to model joint distributions between variables for multivariate forecasting.
	

	\bibliographystyle{IEEEtran}
	\bibliography{QBSD}
	
\begin{appendices}
	
%
%
	
	\section{Experimentation on Popular Datasets}\label{sec:extra}
	
	\subsection{Datasets}
	The evaluation included six open datasets to compare the performance of the proposed method with the state of the art. The following is a description of each dataset with the respective training and testing ranges. Table~\ref{tab:tab1} summarises the statistics of each of these datasets.
	
	Births2015 \cite{births2015} comprises daily records documenting the birth rate observed in the year 2015. A moving training window of one month is employed to predict the next month of data (from 01-02-2015 to 28-02-2015).
	
	Electricity Demand \cite{alex_kozlov_2020}, also known as Daily Electricity Price and Demand Data, contains the electricity demand, price, and weather data in Victoria, Australia, over 2016 days from 01-01-2015 to 06-10-2020. A moving training window of one year is employed to predict one month of data (from 01-01-2016 to 31-01-2016).
	
	Bitcoin Transactional Data \cite{sushil_kumar_2022} contains daily records on the number of Bitcoin transactions from July 2010 to July 2022. A moving training window of one month was employed to predict one year of data (from 01-01-2016 to 31-12-2016).
	
	Electricity \cite{Dua:2019}, also known as the Electricity Load Diagrams dataset from the UCI Machine Learning Repository, contains the electricity load consumption of 321 clients over three years. A moving training window of one year is employed to predict one month of data (from 01-01-2013 00:00:00 to 31-01-2013 23:00:00).
	
	Weather \cite{WTH} provides local climatological data for 1600 U.S. locations over four years from 2010 to 2013 at an hourly interval. A moving training window of one year is employed to predict one week of data (from 01-03-2011 00:00:00 to 07-03-2011 23:00:00).
	
	\subsection{Training procedure}
	As  mentioned before, a moving training window is followed for a single-step forecast for each timestamp for every dataset-algorithm combination. The generic architecture of N-BEATS was utilized for the Births2015 dataset and interpretable architecture for all the other datasets with varying input and output chunk lengths for each dataset. Both N-BEATS and N-HiTS were executed for 100 epochs for Births2015, Electricity Demand, and Bitcoin Transactional datasets, and 50 epochs for Cell-F, Electricity and Weather datasets. The learning rate, n estimators, and max depth for XGBoost and LightGBM were set to 0.01, 1000, and 3, respectively, for all datasets, along with early stopping rounds at 50.
	
	To determine the most accurate forecast for QBSD, the size of the context period $k$ was manually adjusted for each dataset as shown in Table~\ref{tab:tab1}.  

	\begin{table}
	 \small
	 \caption{Dataset statistics and QBSD parameters}
	     \resizebox{\columnwidth}{!}{%
			     \begin{tabular}{lrrrrr}
				 \toprule
				       Dataset & \#Records & Frequency & Window & Target & $k$ \\
				 \midrule
				 Births2015 \cite{births2015} & 365 & daily &  1 month & births & 1 day\\
				 Electricity Demand \cite{alex_kozlov_2020} & 2106 & daily & 1 year & demand  & 2 days \\
				 Bitcoin Transactional \cite{sushil_kumar_2022} & 4389 & daily & 1 month & transactions  & 2 days \\
				 Electricity \cite{Dua:2019} & 17520 & hourly & 1 year & MT\_320 &  2 hours\\
				 Weather  \cite{WTH} & 35064 & hourly & 1 year & WetBulbFarenheit  & 2 hours\\
				 Cell-F 
				 & 8544 & 15 min. & 1 month & KPIs A to F  & 1 hour\\
				 \bottomrule
				     \end{tabular}
			     }
		     \label{tab:tab1}
	
		\vspace{1.1 em}
		\caption{MAPE comparison with open datasets}
		\centering
		\renewcommand{\arraystretch}{1.1}
		\scalebox{0.8}{%
			\begin{tabular}{lrrrrrrc}\toprule
				&\multicolumn{5}{c}{Datasets} \\
				\cmidrule{2-6}
				Method &
				{Births} &
				{Electr. D} &
				{Bitcoin} &
				{Electr.} &
				{Weath.} &
				Mean & p-value$^\dagger$ \\
				\midrule
				Baseline & 14.47 & 9.66 & 8.28 & 3.81 & \textbf{4.40 }& 8.12 & 0.5938 \\
				SARIMA  & 2.85  & 9.33  & 7.06  & 3.75 & 4.61  & 5.52 & 0.9062 \\
				XGBoost  & \textbf{1.11}  & 3.48  &\textbf{ 2.96}  & \textbf{1.02} & 5.38  & \textbf{2.79} & 1.0000 \\
				LightGBM & 18.29 & 7.71  & 9.71  & 4.17 & 11.01 & 10.18 & 0.5938 \\
				Prophet & 2.16  & 8.50  & 6.58  & 6.49 & 29.75 & 10.70 & 0.4062 \\
				N-BEATS & 1.38  & \textbf{0.99}  & 5.14  & 7.23 & 21.35 & 7.22 & 0.6875 \\
				N-HiTS  & 3.16  & 11.35 & 11.04 & 5.64 & 21.18 & 10.47 & 0.0312 \\
				QBSD    & 1.83  & 10.19 & 7.42  & 5.29 & 15.62 & 8.07 & - \\
				\bottomrule
			\end{tabular}}
		\label{tab:mape1}
	  
		\vspace{1.1 em}
		\caption{Execution time$^*$ for training and single prediction}
		\centering
		\scalebox{0.73}{%
			\begin{tabular}{lrrrrrrc}\toprule
				& & & Datasets &\\
				\cmidrule{2-7}
				Method & 
				{Births} &
				{Electr. D} &
				{Bitcoin} &
				{Electr.} &
				{Weath.} &
				Cell-F & p-value$^\dagger$\\
				\midrule
				SARIMA   & 1.03 s & 8.71 s & 278 ms & 5.88 s & 5.91 s  &  2.17 s & \textbf{0.016}\\ 
				XGBoost  &  283 ms & 376 ms & 246 ms & 316 ms & 93.9 ms & 246 ms & \textbf{0.016}\\
				LightGBM  & \textbf{11.1 ms} & 17.3 ms & 15 ms &  88.8 ms & 208 ms & 120 ms & \textbf{0.047}\\
				Prophet &  891 ms & 742 ms & 1.08 s & 4.47 s & 6.81 s & 4.63 s & \textbf{0.016} \\ 
				N-BEATS & 14 s & 4.86 s & 17.7 s & 2 hr 2 s & 800 s & 197 s & \textbf{0.016} \\  
				N-HiTS & 3.34 s & 14.6 s & 14.7 s & 477 s &466 s & 149 s & \textbf{0.016}\\ 
				QBSD  & 14.3 ms & \textbf{13.2 ms} & \textbf{13.6 ms} & \textbf{16.4 ms} & \textbf{21.5 ms} & \textbf{8.72 ms} & -\\
				\bottomrule
			\end{tabular}}
		\label{tab:runtime}
	
		\vspace{1em}\justifying 

		\noindent\begin{itemize}
			\item[$\dagger$]The p-values were obtained through a Wilcoxon signed-rank test comparing the particular method with the QBSD results. The alternative hypotheses posits that QBSD results have lower MAPE and a faster runtime than the corresponding method being tested in Tables \ref{tab:mape1} and \ref{tab:runtime} respectively.
			\item[*] All experiments were conducted through their respective Python implementations on a machine with the following specifications: Intel Core i5-1145G7 @ 2.60 GHz, 16 GB RAM.
		\end{itemize}
	\end{table}

	 \subsection{Results}
	 
	  The following metrics were used to evaluate the performance of each forecasting method: mean absolute error (MAE), mean squared error (MSE), root mean squared error (RMSE), mean absolute percentage error (MAPE), and  $R^2$ statistics.
	  Table~\ref{tab:mape1} presents the MAPE results of these experiments, accompanied by corresponding p-values for statistical validation. Interested readers seeking a comprehensive comparative analysis of the methods across a variety of forecasting metrics can refer to Table~\ref{tab:Moving training window_results} and Table~\ref{tab:EON1-Cell-F_moving training window}. 
	 Table~\ref{tab:runtime} shows the execution time of each algorithm for a single training and prediction cycle. The table shows that QBSD has a statistically shorter runtime than the rest of the methods compared.
	 In the observation considering the KPI dataset, QBSD was over 10 times faster than LightGBM,
	 over 25 times faster than XGBoost, 
	 and over 22,000 times faster than N-BEATS. These algorithms have a time complexity involving a product of multiple variables depending on the hyperparameter configuration. Clearly, QBSD is not the most accurate of forecasters, but is certainly the fastest by a statistically significant margin, i.e., beyond 5\% level of significance $(\text{p-value} < 0.05)$.
	  
	\begin{table}[h!]
		\caption{Comparison of forecasting accuracy with open datasets}
		\addtolength{\tabcolsep}{0pt}
		\centering
			\begin{tabular}{lrrrr}\toprule
				Dataset/method & RMSE & MAE & MAPE & \begin{math}
					R^{2}   
				\end{math} \\ 
				\midrule
				\textit{\textbf{Births2015}} \\
				Baseline M(t-1) & 2082.232 & 1398.5 & 14.471 & -0.258 \\
				Prophet & 301.223 & 228.753 & 2.155 & 0.974 \\ 
				SARIMA & 372.017 & 301.89 & 2.852 & 0.96 \\
				N-BEATS  & 188.366   & 146.338 &  1.384 & 0.990 \\
				XGBoost & 203.673 & 121.515 & 1.11 & 0.988 \\ 
				LightGBM & 1925.7 & 1701.304 & 18.287 & -0.076 \\ 
				N-HiTS  & 396.68 & 332.191 & 3.158 & 0.954 \\ 
				QBSD& 242.485 & 193.304 & 1.83 & 0.983 
				\\  \midrule
				\textit{\textbf{Electricity Demand}} \\
				Baseline M(t-1)  & 15705.789 & 11761.684 & 9.658 & 0.056 \\
				Prophet & 12926.25 & 10606.509 & 8.5 & 0.36 \\ 
				SARIMA & 15202.864 & 11390.745 & 9.329 & 0.115 \\ 
				N-BEATS  & 1574.550   & 1182.520  & 0.986 & 0.991 \\
				XGBoost & 10036.174 & 4841.364 & 3.48 & 0.614 \\ 
				LightGBM & 12195.286 & 9639.455 & 7.707 & 0.431 \\
				N-HiTS & 15442.059 & 13276.156 & 11.346 & 0.087 \\
				QBSD & 15958.825 & 12793.552 & 10.19 & 0.025
				\\ \midrule
				\textit{\textbf{Bitcoin Transactional}} \\
				Baseline M(t-1) & 24671.904 & 18492.888 & 8.282 & 0.474 \\
				Prophet & 19428.508 & 14518.372 & 6.598 & 0.674 \\ 
				SARIMA & 22471.417 & 15897.475 & 7.062 & 0.564 \\ 
				N-BEATS & 22202.616  & 13236.239 & 5.319 & 0.574\\
				XGBoost & 13810.061 & 7102.469 & 2.957 & 0.835 \\
				LightGBM & 26279.781 & 21300.81 & 9.705 & 0.404 \\
				N-HiTS & 32860.864 & 25276.649 & 11.035 &   0.068 \\
				QBSD & 21088.87 & 16285.221 & 7.419 & 0.616  \\
				\midrule
				\textit{\textbf{Electricity}} \\
				Baseline M(t-1) & 180.073 & 115.915 & 3.812 & 0.774 \\
				Prophet & 253.199 & 192.425 & 6.462 & 0.554 \\ 
				SARIMA & 182.245 & 113.967 & 3.748 & 0.769 \\ 
				N-BEATS  & 294.460 & 231.811  & 7.228 & 0.397 \\
				XGBoost & 87.928 & 35.732 & 1.021 & 0.946 \\ 
				LightGBM & 195.901 & 119.781 & 4.171 & 0.733 \\ 
				N-HiTS & 213.884 & 167.355 & 5.639 &  0.682 \\ 
				QBSD & 224.175 & 163.132 & 5.294 & 0.65 
				\\\midrule
				\textit{\textbf{Weather}} \\ 
				Baseline M(t-1) & 1.433 & 0.946 & 4.396 & 0.919 \\
				Prophet & 8.273 & 7.467 & 29.750 & -1.712 \\ 
				SARIMA & 1.456 & 0.997 & 4.610 & 0.915 \\
				N-BEATS &  5.847 & 4.771 & 21.354 & -0.355\\
				XGBoost & 2.314 & 1.472 & 5.381 & 0.787 \\ 
				LightGBM & 3.562 & 1.701 & 11.007 & 0.497 \\
				N-HiTS & 6.009 & 5.281 & 21.176 & -0.431 \\
				QBSD & 4.006 & 3.084 & 15.617 & 0.363 \\
				\bottomrule
			\end{tabular}
		\label{tab:Moving training window_results}
	\end{table}
	
	\begin{table}[h!]
		\caption{Comparison of forecasting accuracy with EON1-Cell-F.}
		\addtolength{\tabcolsep}{0pt}
		\centering
		\renewcommand{\arraystretch}{0.9}
			\begin{tabular}{lrrrr}\toprule
				KPI/Method & RMSE & MAE & MAPE & \begin{math}
					R^{2}   
				\end{math} \\ 
				\midrule
				\textit{\textbf{A}}  \\
				Baseline M(t-1) & 858.952 & 609.96 & 22.23 & 0.841 \\
				Prophet & 1087.553 & 837.236 & 39.839 & 0.727 \\ 
				SARIMA & 832.026 & 599.671 & 19.405 & 0.84 \\ 
				N-BEATS & 607.483 & 467.888 &  15.413 & 0.915 \\
				XGBoost & 504.232 & 242.046 & 5.464 & 0.941 \\
				LightGBM & 650.15 & 358.361 & 14.92 & 0.902 \\
				N-HiTS & 702.113 & 507.212 & 20.566 & 0.893 \\ 
				QBSD & 635.615 & 479.883 & 15.702 & 0.907 
				\\ \midrule
				\textit{\textbf{B}} \\
				Baseline M(t-1) & 2.034 & 1.584 & 23.416 & 0.007 \\
				Prophet & 1.545 & 1.292 & 19.32 & 0.418 \\
				SARIMA & 1.623 & 1.353 & 19.912 & 0.358 \\
				N-BEATS & 1.464 & 1.238 & 18.376 & 0.478 \\
				XGBoost & 1.004 & 0.587 & 6.599 & 0.755 \\
				LightGBM & 1.146 & 0.812 & 12.005 & 0.68 \\
				N-HiTS & 2.032 & 1.589 & 21.867 & 0.008 \\
				QBSD & 1.559 & 1.293 & 18.892 & 0.408 
				\\  \midrule
				\textit{\textbf{C}} \\ 
				Baseline M(t-1) & 149.165 & 106.16 & 24.98 & 0.79 \\
				Prophet & 140.569 & 109.989 & 35.411 & 0.792 \\
				SARIMA & 135.581 & 101.174 & 21.907 & 0.806 \\
				N-BEATS & 107.766  & 83.308 & 18.137 & 0.878 \\
				XGBoost & 80.027 & 39.176 & 5.476 & 0.932 \\ 
				LightGBM & 93.104 & 53.095 & 13.686 & 0.909 \\
				N-HiTS & 117.915 & 91.766 & 31.538 & 0.868 \\
				QBSD & 111.558 & 84.828 & 17.784 & 0.869 
				\\  \midrule
				\textit{\textbf{D}} \\
				Baseline M(t-1) & 181.636 & 138.056 & 54.092 & 0.735 \\
				Prophet & 178.576 & 141.535 & 75.767 & 0.715 \\ 
				SARIMA & 173.143 & 135.051 & 39.819 & 0.732 \\ 
				N-BEATS & 136.521 & 109.982 & 39.902 & 0.833 \\
				XGBoost & 96.582 & 51.005 & 7.16 & 0.917 \\ 
				LightGBM & 115.223 & 73.89 & 38.598 & 0.881\\
				N-HiTS & 149.131 & 114.592 & 52.002 & 0.869 \\
				QBSD & 139.196 & 111.798 & 42.075 & 0.827 
				\\ \midrule
				\textit{\textbf{E}} \\
				Baseline M(t-1) & 8.431 & 6.026 & 7.613 & 0.977 \\
				Prophet & 19.313 & 15.278 & 26.006 & 0.879 \\ 
				SARIMA & 7.697 & 5.656 & 6.878 & 0.981 \\ 
				N-BEATS &  5.948  & 4.593   & 5.735 & 0.989\\
				XGBoost & 4.263 & 2.104 & 2.14 & 0.994 \\ 
				LightGBM & 6.765 & 3.587 & 4.79 & 0.985 \\
				N-HiTS & 7.468 & 5.539 & 7.841 & 0.982 \\
				QBSD & 5.819 & 4.374 & 5.137 & 0.989  
				\\ \midrule
				\textit{\textbf{F}} \\
				Baseline M(t-1) & 5.882 & 3.747 & 99.32 & 0.101 \\
				Prophet & 4.869 & 3.432 & 116.839 & 0.384 \\ 
				SARIMA & 4.954 & 3.183 & 83.598 & 0.362 \\
				N-BEATS & 4.254 & 2.822  & 75.415 &  0.530 \\
				XGBoost & 3.108 & 1.327 & 13.071 & 0.749 \\ 
				LightGBM & 3.492 & 1.937 & 59.796 & 0.683 \\
				N-HiTS & 5.122 & 3.379 & 108.340 & 0.318 \\
				QBSD & 4.415 & 2.886 & 81.881 & 0.494 
				\\\bottomrule
			\end{tabular}
		\label{tab:EON1-Cell-F_moving training window}
	\end{table}

\end{appendices}



\end{document}